\documentclass[conference]{IEEEtran}
\IEEEoverridecommandlockouts
\usepackage{cite}
\usepackage{amsmath,amssymb,amsfonts}
\usepackage{algorithmic}
\usepackage{graphicx}
\usepackage{textcomp}
\usepackage{comment}
\usepackage{hyperref}
\usepackage{tabularx}
\usepackage{microtype}
\usepackage{pifont}% http://ctan.org/pkg/pifont
\newcommand{\xmark}{\ding{55}}%
\usepackage{listings}
\usepackage[table]{xcolor}
\usepackage{multirow}
\usepackage{amsthm}
\theoremstyle{definition}

\usepackage{array}

\newboolean{showcomments}
\setboolean{showcomments}{true} % toggle to show or hide comments
\ifthenelse{\boolean{showcomments}}
  {
		\newcommand{\nbb}[2]{
		\fcolorbox{black}{yellow}{\bfseries\sffamily\scriptsize#1}
		{\sf$\blacktriangleright$\textcolor{blue}{\textit{#2}}$\blacktriangleleft$}
		}
		
		\newcommand{\remarks}[1]{\color{red}[#1]\color{black}}

		\newcommand{\del}[1]{\textcolor{red}{\sout{#1}}} % please delete
  }
  {
		\newcommand{\nbb}[2]{}
		\newcommand{\remarks}[1]{}

		\newcommand{\del}[1]{} % please delete
  }

\def\BibTeX{{\rm B\kern-.05em{\sc i\kern-.025em b}\kern-.08em
    T\kern-.1667em\lower.7ex\hbox{E}\kern-.125emX}}
\begin{document}

\title{Semantic-Aware Representation of Multi-Modal Data for Data Ingress: A Literature Review
\thanks{Funded by Swedish Research Council (VR), Diarienummer: 2023-03810.}}

\author{\IEEEauthorblockN{Pierre Lamart}
\IEEEauthorblockA{\textit{Computer Science and Engineering} \\
\textit{University of Gothenburg}\\
Gothenburg, Sweden \\
pierre.lamart@gu.se}
\and
\IEEEauthorblockN{Yinan Yu}
\IEEEauthorblockA{\textit{Computer Science and Engineering} \\
\textit{Chalmers University of Technology}\\
Gothenburg, Sweden \\
yinan@chalmers.se}
\and
\IEEEauthorblockN{Christian Berger}
\IEEEauthorblockA{\textit{Computer Science and Engineering} \\
\textit{University of Gothenburg}\\
Gothenburg, Sweden \\
christian.berger@gu.se}
}

\maketitle

\begin{abstract}
Machine Learning (ML) is continuously permeating a growing amount of application domains. Generative AI such as Large Language Models (LLMs) also sees broad adoption to process multi-modal data such as text, images, audio, and video. While the trend is to use ever-larger datasets for training, managing this data efficiently has become a significant practical challenge in the industry--double as much data is certainly not double as good. Rather the opposite is important since getting an understanding of the inherent quality and diversity of the underlying data lakes is a growing challenge for application-specific ML as well as for fine-tuning foundation models. Furthermore, information retrieval (IR) from expanding data lakes is complicated by the temporal dimension inherent in time-series data which must be considered to determine its semantic value. This study focuses on the different semantic-aware techniques to extract embeddings from  mono-modal, multi-modal, and cross-modal data to enhance IR capabilities in a growing data lake. Articles were collected to summarize information about the state-of-the-art techniques focusing on applications of embedding for three different categories of data modalities.
\end{abstract}

\begin{IEEEkeywords}
data lake, data modality, multi-modal data, information retrieval, embedding, literature review
\end{IEEEkeywords}

\rowcolors{2}{gray!7}{white}

\section{Introduction}

Facing the growing issue of handling large multi-modal datasets, researchers and practitioners try to enhance data management systems by improving the efficiency of specific functions for data ingress (like computing embeddings from data samples) or identifying relevant data for egress as needed for IR. Understanding how such data can be processed and stored is becoming a critical challenge to benefit from a large volume of data. A growingly popular response for data storage is the use of data lakes that allow to store large volumes of raw data from various sources. However, blindly growing the data lake is unsustainable and it results in barely manageable datasets that are hard to use and investigate (cf.~Udandarao et al., \cite{udandarao2024zeroshot}). Making this approach not sufficient and require proper data pre-processing methods prior to data ingress.

Moreover, the data collected is increasingly multi-modal, including a variety of formats such as text, images, audio, and video. Each modality comes with its own unique type of metadata, requiring distinct organizational structures. This diversity adds another layer of complexity to data lakes, necessitating metadata management solutions that can efficiently index and retrieve data across different modalities based on their semantic content. It is hence valuable to gain insights into the feature extraction and indexing techniques for multi-modal data.

The aim of this review is to provide an overview of existing methods and techniques to encode multi/cross-modal, time-series data in a semantically-aware way. This will allow us to identify possible research trends and get an overview of the state-of-the-art approaches for data ingress. We address the following research question:

\begin{description}
\item[RQ:] What are common approaches to prepare mono/multi/cross-modal data for data ingress, considering the temporal aspects in a growing data lake?
\end{description}

We structure our paper as follows: Sec.~\ref{sec:related_work} introduces relevant concepts and related work. Sec.~\ref{sec:methodology} outlines how we conducted our study. The results are discussed  in Sec.~\ref{sec:results_and_discussion}. The conclusions \& future work are addressed in Sec.~\ref{sec:conclusion}.

%%%%%%%%%%%%%%%%%%%%%%%%%%%%%%%%%%%%%%%%%%%%%%%%%%%%%%%%%%%%%%%%%%%%%%%%%%%%%%%%%%%%%%%%%%%%%%%%%%%%%%%%%%%%%%%%%%%%%%%%%%%%%%%%%%
%%%%%%%%%%%%%%%%%%%%%%%%%%%%%%%%%%%%%%%%%%%%%%%%%%%%%%%%%%%%%%%%%%%%%%%%%%%%%%%%%%%%%%%%%%%%%%%%%%%%%%%%%%%%%%%%%%%%%%%%%%%%%%%%%%

\section{Related Work}
\label{sec:related_work}
Recent technologies and applications relying on Machine Learning (ML)-enabled systems requires an increasing amount of data for continuous improvement. Often coming from different sources, the data is likely to include different modalities \cite{Sukhobokov2022407,Kumari20192152}. Continuously collecting data to benefit from this technology results in data management systems, such as data lakes, becoming insufficient and drowning in the collected data.

Several reviews covering multi-modal systems and applications are available in the literature. Perez-Martin et al.\cite{Perez-Martin20224165} present a comprehensive review of the cross-modal applications and challenges of text and video data. They study the progress of researchers on 26 datasets for “text retrieval from video task and video captioning/description task”. The review highlights that despite all the progress made in that field, there are still many possible improvements to make to extract and describe complex spatiotemporal information within videos.
Similarly, Kaur et al.\cite{Kaur2021} studied cross-modal image-text information retrieval. They compared several approaches to identify their strengths and weaknesses. They see possible improvements in algorithms’ performances. Chen et al.\cite{Chen2021195} conducted a review of deep learning models for the same data modalities. They studied different popular structures for uni-directional and bi-directional multi-modal tasks. Although they see several applications of these models, they argue that adding more modalities to the process could allow these technologies to be applied in more scenarios but is still underexplored.
In 2020 Zhang et al.\cite{Zhang2020478} analyzed work in multi-modal deep learning from three perspectives: “learning multimodal representations, fusing multimodal signals at various levels, and multimodal applications”. The main modalities studied are natural language and computer vision.

Most previous work primarily focuses on techniques using curated datasets, which are typically well-defined for machine learning tasks and benchmarking. However, data within a data lake often has a temporal dimension, as data is continuously collected from various sources. This stresses the importance of the time series data modality, which has not received the same level of attention as other data modalities. In this paper, we highlight the time aspects, focusing on the system capability of handling the dynamic and temporal nature of real-world data.

%%%%%%%%%%%%%%%%%%%%%%%%%%%%%%%%%%%%%%%%%%%%%%%%%%%%%%%%%%%%%%%%%%%%%%%%%%%%%%%%%%%%%%%%%%%%%%%%%%%%%%%%%%%%%%%%%%%%%%%%%%%%%%%%%%
%%%%%%%%%%%%%%%%%%%%%%%%%%%%%%%%%%%%%%%%%%%%%%%%%%%%%%%%%%%%%%%%%%%%%%%%%%%%%%%%%%%%%%%%%%%%%%%%%%%%%%%%%%%%%%%%%%%%%%%%%%%%%%%%%%

\section{Methodology}
\label{sec:methodology}

Our search for articles was inspired by Petersen et al.~(cf.~\cite{mappingstudy}).We focused on articles about data embedding and multi-modal fusion to provide an overview about common ways to embed specific modalities and how these modalities can be fused. Our study follows a similar structure to the one from Zhang et al.\cite{Zhang2020478}. However, we focus on methods that gained attention after 2020, namely, representational and contrastive learning. Representational learning refers to models learning the representation of an input data for a specific task like classification, clustering or, in our case, embedding. Contrastive learning is a recent powerful paradigm that learns to differentiate between similar and dissimilar examples \cite{Kang202381011}. It creates positive and negative pairs, maps the pairs using a non-linear encoder, and optimizes the encoder by minimizing the distance between positive pairs and maximizing the negative pairs \cite{Liu2023}. Contrastive learning gained a lot of attention past 2020. When searching for articles with the keyword “Contrastive Learning” In Scopus, 30 articles were published before 2020 and more than 5000 after 2020 while in ACM Digital Library, all articles were published in 2020 or later.

After piloting our search terms in Scopus, ACM Digital Library, and IEEE xPlore, we refined our query using specific keywords for our subject. The template query for the modality \verb|time-series| is depicted below:

\begin{lstlisting}
    KEY ( representational OR contrastive )
AND KEY ( machine OR deep OR ai )
AND KEY ( learning )
AND TITLE-ABS-KEY ( embedding )
AND TITLE-ABS-KEY ( |\colorbox{blue!10}{time-series}| )
\end{lstlisting}

We systematically replaced the highlighted keyword \verb|time-series| with \verb|image| or \verb|text OR prompt| to find embedding approaches targeting specifically these two data modalities. To limit the result set for the data modality \verb|image|, though, we filtered only on the keywords instead for title, abstract, or keywords.

\begin{table}[h!]
    \centering
    \caption{Number of articles from our initial dataset and after applying our filter criteria.}
    \begin{tabular}{|l|c|c|}
        \rowcolor{gray!50}
        \hline
        & \textbf{Initial Search} & \textbf{Selected Articles} \\
        \hline
        Time-series embedding & 121 & 7 \\
        \hline
        Text embedding & 345 & 2 \\
        \hline
        Image embedding & 1085 & 5 \\
        \hline
        Raw data fusion & 194 & 7 \\
        \hline
        Embedded data fusion & 92 & 4 \\
        \hline
        \hline
        \rowcolor{gray!50}
        \textbf{Total} & \textbf{1837} & \textbf{25} \\
        \hline
    \end{tabular}
    \label{tab:literature_search_left_side}
\end{table}

To find articles focusing on fusing multi-modal raw data, we used the following template query to identify relevant papers:

\begin{lstlisting}
    KEY ( fusion OR
          alignment OR
          coordination OR
          factorization )
AND KEY ( time-series OR text OR image )
AND TITLE-ABS-KEY ( ( machine OR deep )
                      AND learning OR ai
                  )
AND TITLE-ABS-KEY ( modal* OR multi-modal* )
AND TITLE-ABS-KEY ( |\colorbox{blue!10}{raw}| AND data )
\end{lstlisting}

\begin{table*}
    \centering
    \caption{Data embedding approaches targeting mono-modal data.}
    \begin{tabular}{|l|c|l|c|>{\arraybackslash}p{0.38\linewidth}|}
       \rowcolor{gray!50}
        \hline
        \textbf{Model Name} & \textbf{Date} & \textbf{Embedded Data Type} & \textbf{Time Dependency} & \textbf{Embedding Method} \\
        \hline
        COVID-Net (redesign)\cite{Wang20202806} & 2020 & Medical Imaging Data & X & Contrastive Learning \\
        \hline
        Source Model Selection\cite{Cho2021325} & 2021 & Colored Images &  & Supervised Contrastive Learning \\
        \hline
        ReTrim\cite{Garg2021460} & 2021 & Multivariate Time-Series & X & Self-Supervised AutoEncoder + Contrastive Learning \\
        \hline
        CGC\cite{Park20221115} & 2022 & Temporal Graph & X & GNN + Contrastive Learning \\
        \hline
        Pyraformer\cite{Liu2022} & 2022 & Time-Series & X & Embedding Layers \\
        \hline
        CSCL\cite{Tarasiou2022} & 2022 & Annotated Satellite Image &  & Context-Self Contrastive Loss \\
        \hline
        3D FCN\cite{Mohammadi2023272} & 2023 & Time-Series Images & X & Cross-Entropy + Contrastive Learning Supervisions \\
        \hline
        TiCTok\cite{Kang202381011} & 2023 & Multivariate Time-Series & X & Contrastive Learning + Token Encoder \\
        \hline
        SuperConText\cite{Moukafih202316820} & 2023 & Text &  & Neural Network Encoder + Contrastive Learning \\
        \hline
        CoLDE\cite{Jha2023} & 2023 & Long-Form Document &  & Positional Embedding and Attention Layer + Contrastive Learning \\
        \hline
        DTX\cite{Loza202448} & 2024 & Images and Video Frames & X & Transformer-Based with Sel-Attention and Contrastive Loss \\
        \hline
        MabCUT\cite{Yang2024} & 2024 & Images &  & Separate Embedding Blocks + Contrastive Learning \\
        \hline
        PDE\cite{Xu2024} & 2024 & Time-Series Images & X & Contrastive Learning \\
        \hline
    \end{tabular}
    \label{tab:mono-modal_data}
\end{table*}

\begin{table*}[h!]
    \centering
    \caption{Approaches for data embedding for multi-modal data including fusion}
    \begin{tabular}{|l|c|>{\arraybackslash}p{0.15\linewidth}|c|l|>{\arraybackslash}p{0.19\linewidth}|}
       \rowcolor{gray!50}
        \hline
        \textbf{Model Name} & \textbf{Date} & \textbf{Fused Data Types} & \textbf{Time Dependency} & \textbf{Fusion Domain} & \textbf{Fusion Method} \\
        \hline
        CRF-Net\cite{Nobis2019} & 2019 & Camera and Radar Data & X & Raw Data & Concatenation fed into a convolutional network \\
        \hline
        FCN\cite{Caltagirone2019125} & 2019 & Camera Images and LIDAR Point Cloud & X & Cross Fusion & Cross Fusion Fully Convolutional Network \\
        \hline
        GroupFusionNet\cite{Cao202158} & 2021 & fundus image, visual field tests and age &  & Early and Late Fusions & 2 ResNet to Fuse Features \\
        \hline
        Text Image Residual Gating\cite{Sarker20221543} & 2022 & Images and Text &  & Late Fusion & Feature Fusion Layer \\
        \hline
        TGDT\cite{Liu20233622} & 2023 & Images and Text &  & Late Fusion & Transformer-Based Feature Extraction \\
        \hline
        MM-FI\cite{Yang2023} & 2023 & RGB-D Frames, Point Cloud and WiFi CSI Data & X & Raw Data & Least Mean Square Algorithm \\
        \hline
        TF-YOLO\cite{Chen2023} & 2023 & Visible and Infrared Images & X & Adaptative Fusion & Transformer-Fusion Module in a Backbone Network \\
        \hline
        ITContrast\cite{Wu2024} & 2024 & Images and Text &  & Embedded Data Fusion & Contrastive Learning for Image-Text Matching \\
        \hline
        MCSTransWnet\cite{Cheng2024} & 2024 & 3D Topography and Surgical Parameters &  & Late Fusion & Transformer Model and CNN Model \\
        \hline
        CVG Classification\cite{Narotamo2024} & 2024 & Signals and Images & X & Early, Late, and Joint & CNN-RNN Models \\
        \hline
    \end{tabular}
    \label{tab:multi-modal_data}
\end{table*}

We replaced the highlighted keyword \verb|raw| by \verb|embedded| to search for other type of fusion. We continued our selection process by screening the articles' title and abstract by applying the following inclusion/exclusion criteria:

\begin{itemize}
\item[$\checkmark$] Paper addressing the research question.
\vspace{0.2cm}
\item[\xmark] Non-peer-reviewed articles
\item[\xmark] Publications for which full text is not available.
\item[\xmark] Articles written in an other language than English
\item[\xmark] Duplicate papers and shorter versions of already included publications.
\end{itemize}
Tab.~\ref{tab:literature_search_left_side} summarizes how the search results were narrowed down after applying our inclusion/exclusion criteria.

\noindent{\bf Threats to Validity:}
We report potential threats to the validity of our research following recommendations by Feldt and Magazinius (cf.~\cite{validity-threats}). A potential threat may originate by the internal design or construction of our study, in particular the literature review that may be prone to subjective selection bias with respect to the identified research papers. We aimed to mitigate this threat by documenting transparently the search terms, databases, and filtering criteria. While potential papers may have been filtered out, our resulting tables (cf.~Tab.~\ref{tab:mono-modal_data} and \ref{tab:multi-modal_data}) document major aspects that we have encountered while screening relevant literature. With respect to generalizability of our findings, we state that we searched for relevant research in a way that is agnostic to a particular application domain. However, specific domains may have certain, domain-relevant constraints that may render generic solutions non-applicable. Such constraints, though, are left for future and domain-specific studies.

%%%%%%%%%%%%%%%%%%%%%%%%%%%%%%%%%%%%%%%%%%%%%%%%%%%%%%%%%%%%%%%%%%%%%%%%%%%%%%%%%%%%%%%%%%%%%%%%%%%%%%%%%%%%%%%%%%%%%%%%%%%%%%%%%%
%%%%%%%%%%%%%%%%%%%%%%%%%%%%%%%%%%%%%%%%%%%%%%%%%%%%%%%%%%%%%%%%%%%%%%%%%%%%%%%%%%%%%%%%%%%%%%%%%%%%%%%%%%%%%%%%%%%%%%%%%%%%%%%%%%

\section{Results and Discussion}
\label{sec:results_and_discussion}
Embedding is a widely adopted semantic encoding technique and hence, the literature review focuses on articles about data embedding and multi-modal fusion.

As can be seen in Tab.~\ref{tab:mono-modal_data}, the majority of embedding techniques rely on contrastive learning. Used with different models, the contrastive approach allows for a better semantic representation of the embedded data. Several researchers like Wang et al.\cite{Wang20202806} or Cho et al.\cite{Cho2021325} use the benefits of both embeddings and contrastive learning for efficient classification. These approaches show better results than others before, however, studies like the ones from Liu et at.~\cite{Liu2023} or Xu et al.~\cite{Xu2024} also highlight the lack of scalability to multi-modal data and the need to push research in this direction.

To understand how multi-modal data can be handled, we looked into different fusion techniques for embedding several modalities. Tab.~\ref{tab:multi-modal_data} shows that most of the collected articles use early or late fusions and 3 use fusions at a variable stage of the process. Cao et al.~\cite{Cao202158} highlight that while early fusion usually outperforms late fusion thanks to low-level features fusion, making it easier to learn using a large amount of data, late fusion allows for a more detailed feature extraction before fusion. Their model takes advantage of both approaches by fusing features by groups in both early and late stages. Caltagirone et al.~\cite{Caltagirone2019125} propose a model optimizing the position of the fusion during the training phase for better results. Chen et al.~\cite{Chen2023} push the flexibility further by implementing a dynamic fusion process that adapts itself according to the input. 

%%%%%%%%%%%%%%%%%%%%%%%%%%%%%%%%%%%%%%%%%%%%%%%%%%%%%%%%%%%%%%%%%%%%%%%%%%%%%%%%%%%%%%%%%%%%%%%%%%%%%%%%%%%%%%%%%%%%%%%%%%%%%%%%%%
%%%%%%%%%%%%%%%%%%%%%%%%%%%%%%%%%%%%%%%%%%%%%%%%%%%%%%%%%%%%%%%%%%%%%%%%%%%%%%%%%%%%%%%%%%%%%%%%%%%%%%%%%%%%%%%%%%%%%%%%%%%%%%%%%%

\section{Conclusions \& Future Work}
\label{sec:conclusion}

Efficient information retrieval from growing data lakes is essential as it allows querying the data lake for semantically relevant information of interest. Furthermore, fast and accurate query processing based on semantic-aware data embedding enables similarity look-ups or determining what information is \emph{not} present yet in a data lake. 
Our review of recent literature covers a critical part of data curation: data ingress,i.e., the preparation of data samples before storing them in a data lake, by screening state-of-the-art approaches.
This process faces challenges from the type and nature of the various data modalities that need to be efficiently handled in data lakes. While we have already surpassed single modalities, today's challenges originate from handling not only multi-modal data but also their data samples over time. 
More specifically, our survey has unveiled the recent dominance of constrastive learning techniques in embedding mono-modal data as they are capable of effectively capturing semantic representations across various data types. Temporal aspects are being increasingly integrated into embedding methods to address the dynamic nature of data collection. Fusion techniques for multi-modal data primarily use early and late fusion, while innovative methods such as dynamic and adaptive fusion have been developed to balance flexibility and efficiency in handling the complexity and diversity of multi-modal data, which is an important direction for future research.

\section*{Acknowledgments}
This research is funded by the Swedish Research Council (Diarienummer: 2024-2028).

%%%%%%%%%%%%%%%%%%%%%%%%%%%%%%%%%%%%%%%%%%%%%%%%%%%%%%%%%%%%%%%%%%%%%%%%%%%%%%%%%%%%%%%%%%%%%%%%%%%%%%%%%%%%%%%%%%%%%%%%%%%%%%%%%%
%%%%%%%%%%%%%%%%%%%%%%%%%%%%%%%%%%%%%%%%%%%%%%%%%%%%%%%%%%%%%%%%%%%%%%%%%%%%%%%%%%%%%%%%%%%%%%%%%%%%%%%%%%%%%%%%%%%%%%%%%%%%%%%%%%

\bibliography{references} 
\bibliographystyle{ieeetr}

%%%%%%%%%%%%%%%%%%%%%%%%%%%%%%%%%%%%%%%%%%%%%%%%%%%%%%%%%%%%%%%%%%%%%%%%%%%%%%%%%%%%%%%%%%%%%%%%%%%%%%%%%%%%%%%%%%%%%%%%%%%%%%%%%%
%%%%%%%%%%%%%%%%%%%%%%%%%%%%%%%%%%%%%%%%%%%%%%%%%%%%%%%%%%%%%%%%%%%%%%%%%%%%%%%%%%%%%%%%%%%%%%%%%%%%%%%%%%%%%%%%%%%%%%%%%%%%%%%%%%
\end{document}